\documentclass[11pt,a4paper]{article}
\usepackage[hyperref]{acl2019}
\usepackage{times}
\usepackage{latexsym}
\usepackage{graphicx}  
\usepackage{multirow}
\usepackage{makecell}
\usepackage{tikz}
\usepackage{tikz-dependency}
\usepackage[inline]{enumitem}
\graphicspath{ {./image/} }
\usepackage{amssymb}
\usepackage{xcolor}
\usepackage{enumitem}
\usepackage{url}

\newcommand{\fullconstituency}{\textbf{Full-C}}
\newcommand{\srlconstituency}{\textbf{SRL-C}}
\newcommand{\dep}{\textbf{Dep}}

\newcommand{\conll}[1]{\textbf{CoNLL'{#1}}}

\newcommand{\minus}{\scalebox{0.75}[1.0]{$-$}}

\aclfinalcopy 
\def\aclpaperid{592} 

\title{How to best use Syntax in Semantic Role Labelling}

\author{Yufei Wang$^{1}$ \and Mark Johnson$^{1}$ \and Stephen Wan$^{2}$ \and Yifang Sun$^{3}$ \and Wei Wang$^{3}$ \\
Macquarie University, Sydney, Australia$^1$ \\
CSIRO Data61, Sydney, Australia$^{2}$ \\
The University of New South Wales, Sydney, Australia$^{3}$ \\
\texttt{yufei.wang@students.mq.edu.au, mark.johnson@mq.edu.au} \\
\texttt{stephen.wan@data61.csiro.au} \\
\texttt{\{yifangs,weiw\}@cse.unsw.edu.au} \\
}

\date{}

\begin{document}
\maketitle

\begin{abstract}


There are many different ways in which external information might be used in an NLP task.  This paper investigates how external syntactic information can be used most effectively in the Semantic Role Labeling (SRL) task.  We evaluate three different ways of encoding syntactic parses and three different ways of injecting them into a state-of-the-art neural ELMo-based SRL sequence labelling model. We show that using a constituency representation as input features improves performance the most, achieving a new state-of-the-art for non-ensemble SRL models on the in-domain \conll{05} and \conll{12} benchmarks.\footnote{Our model source code is available in \url{https://github.com/GaryYufei/bestParseSRL}}


\end{abstract}

\section{Introduction}

Properly integrating external information into neural networks has received increasing attention recently~\cite{D18-1310,P17-1064,D18-1548}. Previous research on this topic can be roughly categorized into three classes: \textbf{i)} \textbf{Input}: The external information are presented as additional input features (i.e., dense real-valued vectors) to the neural network~\cite{collobert2011natural}. \textbf{ii)} \textbf{Output}: The neural network is trained to predict the main task and the external information in a multi-task approach~\cite{C18-1251}.  \textbf{iii)} \textbf{Auto-encoder}: This approach, recently proposed by~\citet{D18-1310}, simultaneously combines the \textbf{Input} and \textbf{Output} during neural models training. The simplicity of these methods allow them to apply to many NLP sequence tasks and various neural model architectures. 

However, previous studies often focus on integrating word-level shallow features such as POS or chunk tags into the sequence labelling tasks. Syntactic information, which encodes the long-range dependencies and global sentence structure, has not been studied as carefully. This paper fills this gap by integrating syntactic information to the sequence labelling task. We address three questions: 
\textbf{1)} \emph{How should syntactic information be encoded as word-level features?} \textbf{2)} \emph{What is the best way of integrating syntactic information?} and \textbf{3)} \emph{What effect does the choice of syntactic representation have on the performance?} 

We study these questions in the context of Semantic Role Labelling (SRL). A SRL system extracts the predicate-argument structure of a sentence.\footnote{\emph{who} did \emph{what} to \emph{whom}, \emph{where} and \emph{when}} Syntax was an essential component of early SRL systems~\cite{xue2004calibrating,J08-2005}. The state-of-the-art neural SRL systems use a neural sequence labelling model without any syntax knowledge~\cite{P18-2058,P17-1044,tan2018deep}. We show below that injecting external syntactic knowledge into a neural SRL sequence labelling model can improve the performance, and our best model sets a new state-of-the-art for a non-ensemble SRL system.

In this paper we express the external syntactic information as vectors of discrete features, because this enables us to explore different ways of injecting the syntactic information into the neural SRL model. Specifically, we propose three different syntax encoding methods: \textbf{a)} a full constituency tree representation (\fullconstituency{}); \textbf{b)} an SRL-specific span representation (\srlconstituency{}); and \textbf{c)} a dependency tree representation (\dep{}). For \textbf{(a)} we adapt the constituency parsing representation from~\cite{D18-1162} and encode the tree structure as a set of features for word pairs. For \textbf{(b)}, we use a categorical representation of the constituency spans that are most relevant to SRL tasks based on~\cite{xue2004calibrating}. Finally, \textbf{(c)} we propose a discrete vector representation that encodes the head-modifier relationships in the dependency trees.

We evaluate the effectiveness of these encodings using three different integration methods on the SRL \conll{05} and \conll{12} benchmarks. We show that using either of the constituency representations in either the \textbf{Input} or the \textbf{Auto-Encoder} configurations produces the best performance. These results are noticeably better than a strong baseline and set a new state-of-the-art for non-ensemble SRL systems.

\section{Related Work}
Semantic Role Labeling (SRL) generally refers to the PropBank style of annotation~\cite{palmer-etal-2005-proposition}. Broadly speaking, prior work on SRL makes use of syntactic information in two different ways. ~\citet{carreras2005introduction,pradhan2013towards} incorporate constituent-structure span-based information, while ~\citet{hajic-etal-2009-conll} incorporate dependency-structure information.

This information can be incorporated into an SRL system in several different ways. ~\citet{swayamdipta2018syntactic} use span information from constituency parse trees as an additional training target in a multi-task learning approach, similar to one of the approaches we evaluate here. ~\citet{roth-lapata-2016-neural} use an LSTM model to represent the dependency paths between predicates and arguments and feed the output as the input features to their SRL system. ~\citet{marcheggiani-titov-2017-encoding} use Graph Convolutional Network~\cite{DBLP:conf/icml/NiepertAK16} to encode the dependency parsing trees into their LSTM-based SRL system. ~\citet{AAAI7468} represent dependency parses using position-based categorical features of tree structures in a neural model.  ~\citet{D18-1548} use dependency trees as a supervision signal to train one of attention heads in a self-attentive neural model. 

\section{Syntactic Representation}
This section introduces our representations of constituency and dependency syntax trees.

\begin{figure}[htbp]
\centering
\includegraphics[width=0.48\textwidth]{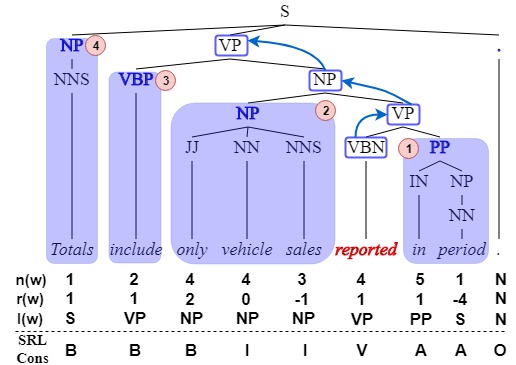}
\caption{Examples of \fullconstituency{} ($n(w)$, $r(w)$ and $l(w)$) and \srlconstituency{} (SRL-Cons). \textit{reported} is the predicate word. The blue non-terminals are candidate constituents in the \srlconstituency{}. The circled number is the extraction order.}
\label{cons}
\end{figure}

\subsection{\fullconstituency{}: Full Constituency Representation}
~\citet{D18-1162} propose a full representation of constituency parsing trees where the string position between $w_i$ and $w_{i+1}$ is associated with the pair $(n(w_{i}) - n(w_{i-1}), l(w_i))$ where $n(w_i)$ is the number of common ancestors between $(w_i, w_{i+1})$ and $l(w_i)$ is the non-terminal label at the lowest common ancestor\footnote{The full constituency trees can be reconstructed from this representation, details refer to~\cite{D18-1162}}. For simplicity, we define $r(w_i) = n(w_{i}) - n(w_{i-1})$ throughout this paper. \footnote{In~\cite{D18-1162}, both $r(w_i)$ and $n(w_i)$ is applicable for this encoding method. Our pilot experiments show that $r(w_i)$ works much better than the absolute representation $n(w_{i})$.}

This encoding method transforms the whole constituency parsing tree into $n \minus 1$ $(r(w_i), l(w_i))$ feature pairs for a length-$n$ sentence. We assign $(r(w_i), l(w_i))$ to the $w_i$ ($0 < i \leq n \minus 1$) and leave a padding symbol \textbf{N} to the $w_n$. We treat $r(w_i)$ and $l(w_i)$ as two separate categorical features for each word. We refer this representation as the \fullconstituency{} (Figure \ref{cons}).

\subsection{\srlconstituency{}: SRL Span Representation}
\citet{xue2004calibrating} show only a small fraction of the constituents in the parse trees are useful for the SRL task given the predicate word. That means encoding the full constituency parsing tree may introduce redundant information. 

Therefore, we preserve the constituent spans that are most likely to be useful for the predicate word in the trees. We re-use the \textit{pruning algorithm} in~\cite{xue2004calibrating}. Their algorithm collects the potential argument constituents by walking up the tree to the root node recursively, which filters out many irrelevant constituents from the syntax trees with 99.3\% of the ground truth arguments preserved. 

We encode the output of this rule-based  pruning algorithm using a standard \textbf{BIO} (\textbf{B}egin-\textbf{I}nside-\textbf{O}utside) annotation scheme. The words that are outside any candidate constituent receive the tag \textbf{O}. The words that are beginning of a candidate constituent receive the tag \textbf{B}, and the words that are inside a candidate constituent receive the tag \textbf{I}.  We use the tag \textbf{A} to label words in prepositional phrases. We refer this representation as the \srlconstituency{} (Figure \ref{cons}).  

\subsection{\dep{}: Dependency Tree Representation}

\begin{figure}[htbp]
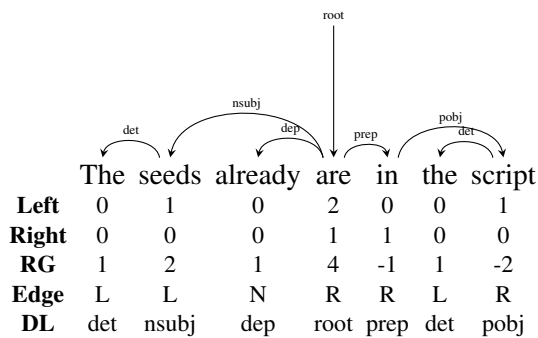

\centering

\begin{dependency}[theme = simple]
\begin{deptext}[font=\small]
\textbf{} \& \normalsize The \& \normalsize seeds \&\normalsize already \&\normalsize are \& \normalsize in \&\normalsize the \&\normalsize script \\
\textbf{Left} \& 0 \& 1 \& 0 \& 2 \& 0 \& 0 \& 1 \\ 
\textbf{Right} \& 0 \& 0 \& 0 \& 1 \& 1 \& 0 \& 0 \\
\textbf{RG} \&1 \& 2 \& 1 \& 4 \& -1 \& 1 \& -2 \\
\textbf{Edge} \& L \& L \& N \& R \& R \& L \& R \\
\textbf{DL} \& det \& nsubj \& dep \& root \& prep \& det \& pobj \\
\end{deptext}
\deproot{5}{root}
\depedge{3}{2}{det}
\depedge{5}{3}{nsubj}
\depedge{5}{4}{dep}
\depedge{5}{6}{prep}
\depedge{8}{7}{det}
\depedge{6}{8}{pobj}
\end{dependency}

\caption{Features from Dependency Tree.}
\label{dep}
\end{figure}

The dependency tree representation encodes key aspects of the head-modifier relationships within the sentence. We also consider encoding constituent edge information. The following word-level features have been proposed:

\begin{enumerate}[label={\alph*)},itemsep=0pt]
\item \emph{\#left/right Dependents} (\textbf{Left} / \textbf{Right}). The number of dependents a word has on the left and right side.
\item \emph{Right/Left-most Dependent} (\textbf{Edge}). Whether the word is the Right/Left/None-most dependent of its governor.
\item \emph{Relative Distance to Governor} (\textbf{RG}). The relative distance between the word and its governor.
\item \emph{Dependency Label} (\textbf{DL}). The label describing the relationship between each pair of dependent and governor.
 
\end{enumerate}
We refer this representation as the \dep{} (Figure \ref{dep}\footnote{In this example, we assume the ``root'' is the first word of the sentence from the left.}).

\section{Injecting External Information}
In this section, we introduce three different methods for integrating external syntactic information into the neural SRL system (Figure \ref{model}):

\begin{figure}[htbp]
\centering
\includegraphics[width=0.4\textwidth]{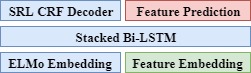}
\caption{Model Architecture. Blue indicates the baseline model; Red indicates the multi-task output component; Green indicates the external feature component.}
\label{model}
\end{figure}

\paragraph{Baseline} Our baseline system is a stacked bi-LSTM architecture~\cite{P17-1044}. We use ELMo~\cite{N18-1202} as word embeddings and a CRF output decoder on the top of LSTM, as shown in Figure \ref{model}. 

\paragraph{Input} This approach represents the external categorical features as trainable, high dimensional dense vector token embeddings, which are concatenated with the representation vectors of ELMo in the baseline model. The syntactic parse trees that are used as the input features are produced by~\citet{Kitaev-2018-SelfAttentive} (for constituency parsing). The dependency trees are produced by transforming the constituency trees using Stanford CoreNLP toolkit. This ensures that the constituency and dependency parses have a similar error distribution, helping to control for parsing quality. Our constituency and dependency parses score a state-of-the-art 95.4 F1 and 96.4\% UAS on the WSJ test set respectively. We used a 20-fold cross-validation procedure to produce the data for the external syntactic input. 

\paragraph{Output} In this approach, our model predicts both SRL sequence tags and syntactic features (encoded as the word-level features above) simultaneously. We use a log loss for each categorical feature. The final training loss is the multi-task objective $L_{SRL} - \sum_{f=1}^m \log p_f(y^\star_f)$, where $p_f(y_f)$ is the probability of generating $y_f$ as the $f^{th}$ feature ($m$ features in total, $m = 1, 2, 5$ for \srlconstituency{}, \fullconstituency{} and \dep{} respectively) and $y^\star_f$ is the ground truth for the $f^{th}$ feature. Gold training data was used as the external syntactic information for the multi-task output setting, as this external information is not required at test time.


\paragraph{Auto-encoder} Following~\citet{D18-1310}, we use external information as input features and as a multi-task training objective simultaneously, so the system is behaving somewhat like an auto-encoder. This auto-encoder has to reproduce the syntactic information in its output that it is fed in its input, encouraging it to incorporate this information in its internal representations. The input and output representations are the same as above. 


\section{Experiments}
We evaluate 10 different models (the 3 ways of using external information by 3 different encodings of syntax and a baseline model) on \conll{05}~\cite{carreras2005introduction} and \conll{12}~\cite{pradhan2013towards} benchmarks, under the evaluation setting where the gold predicate is given. The \conll{05} benchmark uses WSJ and Brown test as in-domain and out-domain evaluation respectively. 

\subsection{Main Results}
Table \ref{ifot} shows the effect of using the three different kinds of external syntactic information in the three different ways just described. When used as input features, all three representations improve over our baseline system. This shows that syntactic representations provide additional useful information, which is beyond the dynamic context embeddings from ELMo, to SRL task.

\paragraph{Syntax Representations}
Models using constituency representations are 0.3\% - 0.6\% better than the models using the dependency representations. This might be because constituents align more directly with SRL arguments and constituency information is easier to use. 

\begin{table}[htbp]
\begin{center}
\begin{tabular}{|c|c|c c|c|}
\hline
\multirow{2}{*}{\small \bf Inject.} & \multirow{2}{*}{\bf \small Model} & \multicolumn{2}{c|}{\small \conll{05}} & \small \conll{12} \\ 
& & \small WSJ & \small Brown & \small Test \\
\hline
- & \small Baseline & 87.7 & 78.1 & 85.8 \\
\hline
\multirow{2}{*}{\small Input} & \small \fullconstituency & 88.1 & 78.9 & \textbf{86.4}  \\
& \small \srlconstituency & \textbf{88.2} & \textbf{79.3} &  \textbf{86.4} \\
& \small \dep & 87.9 & 78.4 & 86.1  \\
\hline
\multirow{2}{*}{\small Output} & \small \fullconstituency & 87.7 & 78.4 & 85.9  \\
& \small \srlconstituency & 87.9 & 78.5 & 85.9  \\
& \small \dep & 87.6 & 78.9 & 85.8  \\
\hline
\multirow{2}{*}{\makecell{\small Auto \\ \small Encoder}} &\small  \fullconstituency & \textbf{88.2} & 77.7 & 86.3  \\
& \small \srlconstituency & \textbf{88.2} & 79.0 & \textbf{86.4}  \\
& \small \dep & 87.6 & 78.1 & 85.7  \\
\hline
\end{tabular}
\end{center}
\caption{\label{ifot} Injecting External Syntax Information. \textbf{Bold number} is the best performance in each column, same below.}
\end{table}

The \srlconstituency{} is slightly better than the \fullconstituency{} for in-domain evaluation. The advantages of the \srlconstituency{} approach are greater on the out-of-domain (Brown) evaluation, with a margin of 0.4\%. This could be because \fullconstituency{} is more sensitive to parsing errors than \srlconstituency{}. When we compare gold and automatic parser representations in Brown device data, 10.5\% of the words get different \fullconstituency{} features while this only 7.9\% get different \srlconstituency{} features.

\paragraph{External Information Injection}
Table \ref{ifot} shows at least on this task, multi-task learning does not perform as well as adding external information as additional input features. Both the \textit{Input} and \textit{Auto-Encoder} methods work equally well. We conclude that the extra complexity of the \textit{auto-encoder} model is not justified. In particular, \dep{} with \textit{auto-encoder} hurts SRL accuracy (0.6\% behind the model with the constituency features).

\subsection{Comparison with existing systems}
We compare our best system (\srlconstituency{} used as Input) with previous work in Table \ref{comparsion}. We improve upon the state-of-the-art results for non-ensemble SRL models on in-domain test by 0.6\% and 0.2\% on \conll{05} and \conll{12} respectively. Our model also achieves a competitive result on \conll{05} Brown Test. Comparing with the strong ensemble model in~\cite{D18-1191}, our model is only 0.3\% and 0.6\% lower in two benchmarks respectively.

\begin{table}[htbp]
\begin{center}
\begin{tabular}{|c|c c|c|}
\hline 
\multirow{2}{*}{\small \bf Model} & \multicolumn{2}{c|}{\small \conll{05}} & \small \conll{12} \\
&\small WSJ & \small Brown & \small Test \\ 
\hline
\small ELMo Baseline & 87.7 & 78.1 & 85.8 \\
\small \citet{D18-1548} & 86.0 & 76.5 & - \\
\small \citet{AAAI7468} & 86.9 & 76.8 & - \\
\small \citet{P18-2058} & 87.4 & \textbf{80.4} & 85.5 \\
\small \citet{D18-1191} & 87.6  & 78.7 & 86.2 \\
\small Our best model & \textbf{88.2} & 79.3 & \textbf{86.4} \\
\hline
\hline
\small \citet{AAAI7468}$^{\S}$ & 87.8 & 78.8 & - \\ 
\small \citet{D18-1191}$^{\S}$ & 88.5 & 79.6 & 87.0 \\ 
\hline
\end{tabular}
\end{center}
\caption{\label{comparsion} Comparison with existing systems. $^{\S}$ indicates ensemble models.}
\end{table}

\subsection{Using Gold Parse Trees}

Finally, we conduct an oracle experiment where all syntactic features are derived from gold trees. Our model performance improves by around 3\% - 4\% F1 score (see Table \ref{golden}).  This bounds the improvement in SRL that one can expect with improved syntactic parses.

\begin{table}[htbp]
\begin{center}
\begin{tabular}{|c|c c|c|}
\hline 
 \multirow{2}{*}{\small \bf Model} & \multicolumn{2}{c|}{\small \conll{05}} &\small \conll{12} \\
& \small WSJ & \small Brown & \small Test \\
\hline
\small Our best model & 88.2 & 79.3 & 86.4 \\
\hline
\small \fullconstituency & \textbf{92.2} & \textbf{83.5} & \textbf{91.4}  \\
\small \srlconstituency & 91.7 & 83.4 & 90.3  \\
\small \dep & 91.9 & 83.3 & 91.1  \\
\hline
\end{tabular}
\end{center}
\caption{\label{golden} SRL Performance with Gold Trees}
\end{table}
\section{Conclusion and Future Work}


This paper evaluated three different ways of representing external syntactic parses, and three different ways of injecting that information into a state-of-the-art SRL system.  We showed that representing the external syntactic information as constituents was most effective.  Using the external syntactic information as input features was far more effective than a multi-task learning approach, and just as effective as an auto-encoder approach. Our best system sets a new state-of-the-art for non-ensemble SRL systems on in-domain data.  

In future work we will explore how external information is best used in ensembles of models for SRL and other tasks.  For example, is it better for all the models in an ensemble to use the same external information, or is it more effective if they make use of different kinds of information?  We will also investigate whether the choice of method for injecting external information has the same impact on other NLP tasks as it does on SRL.


\section*{Acknowledgments}
This research was supported by the Australian Research Council’s Discovery Projects funding scheme (project number DPs 160102156, 170103710, 180103411), D2DCRC (DC25002, DC25003), and in part by CSIRO Data61. 

\bibliographystyle{acl_natbib}
\bibliography{citation}

\end{document}